\documentclass[runningheads]{llncs}

\usepackage{amsmath,amssymb,amsfonts}
\usepackage{graphicx}
\usepackage{booktabs}
\usepackage{multirow}
\usepackage{bm}
\usepackage{float}
\usepackage{placeins}
\usepackage[pagebackref,breaklinks,colorlinks,citecolor=blue]{hyperref}
\usepackage{cleveref}

\begin{document}
\emergencystretch=1.5em

\title{PCReg-Net: Progressive Contrast-Guided Registration for Cross-Domain Image Alignment}
\titlerunning{PCReg-Net: Progressive Contrast Registration}

\author{Jiahao Qin\textsuperscript{*}}
\authorrunning{J. Qin}
\institute{Email: jiahao.qin19@gmail.com}

\maketitle
\let\thefootnote\relax\footnotetext{* Corresponding author.}

\begin{abstract}
\emergencystretch=1em
Deformable image registration across heterogeneous domains remains challenging because coupled appearance variation and geometric misalignment violate the brightness constancy assumption underlying conventional methods. We propose PCReg-Net, a progressive contrast-guided registration framework that performs coarse-to-fine alignment through four lightweight modules: (1)~a registration U-Net for initial coarse alignment, (2)~a reference feature extractor capturing multi-scale structural cues from the fixed image, (3)~a multi-scale contrast module that identifies residual misalignment by comparing coarse-registered and reference features, and (4)~a refinement U-Net with feature injection that produces the final high-fidelity output. We evaluate on the FIRE-Reg-256 retinal fundus benchmark, demonstrating improvements over both traditional and deep learning baselines. Additional experiments on two microscopy benchmarks further confirm cross-domain applicability. With only 2.56M parameters, PCReg-Net achieves real-time inference at 141\,FPS.
Code is available at \url{https://github.com/JiahaoQin/PCReg-Net}.

\end{abstract}

\section{Introduction}
\label{sec:introduction}

Deformable image registration---aligning a moving image to a fixed reference---is a core task across computer vision and biomedical imaging~\cite{chen2024survey,sotiras2013deformable}. Applications span retinal fundus alignment~\cite{hernandez2017fire} and microscopy imaging~\cite{zhang2026orpamreg4k}, each presenting distinct deformation regimes and domain-shift characteristics. A robust registration method must generalize across these diverse scenarios.

Classical registration methods, including SIFT~\cite{lowe2004distinctive}, Demons~\cite{vercauteren2009diffeomorphic}, optical flow~\cite{horn1981determining}, and SyN~\cite{avants2008symmetric}, attempt direct spatial alignment through explicit correspondence estimation or iterative optimization. However, these approaches assume comparable intensity distributions between source and target images, limiting their effectiveness when appearance variation accompanies geometric misalignment~\cite{chen2024survey}. Recent deep learning methods such as VoxelMorph~\cite{balakrishnan2019voxelmorph} and TransMorph~\cite{chen2021transmorph} learn to predict deformation fields from image pairs but similarly rely on intensity similarity, yielding suboptimal results under cross-domain conditions. Scene-appearance separation frameworks~\cite{qin2026domainreg,qin2026sasnet} address domain shift through disentangled representations, but the generative architecture introduces reconstruction noise that limits fine-grained alignment fidelity.

We observe that the key limitation of prior methods is the indirect treatment of alignment: deformation-based approaches warp pixel intensities without accounting for appearance variation, while generative approaches address appearance differences but introduce reconstruction noise. In contrast, a direct image-to-image registration paradigm that progressively refines alignment through explicit comparison with the reference image can circumvent both limitations.

In this paper, we propose PCReg-Net, a progressive contrast-guided registration framework for cross-domain image alignment. The core idea is to decompose registration into two stages: a coarse alignment that approximates the target, followed by a contrast-guided refinement that identifies and corrects residual discrepancies by explicitly comparing coarse-registered features with reference features at multiple scales. The main contributions are:

\begin{enumerate}
    \item We propose PCReg-Net, a progressive coarse-to-fine registration framework consisting of four lightweight modules (2.56M parameters) that achieves high-fidelity alignment by separating coarse registration from contrast-guided refinement.

    \item We introduce a multi-scale contrast module that generates residual alignment cues by comparing features from the coarse-registered and reference images, along with a feature injection mechanism that guides the refinement network using these contrast signals.

    \item We conduct comprehensive evaluation on FIRE-Reg-256 (retinal fundus) with comprehensive baseline comparison, and demonstrate cross-domain applicability on two additional microscopy benchmarks.
\end{enumerate}

\section{Related Work}
\label{sec:related}

\paragraph{Classical registration.}
Feature-based methods such as SIFT~\cite{lowe2004distinctive} estimate sparse correspondences and recover geometric transformations, performing well for rigid alignment with sufficient texture but degrading under appearance variation. Intensity-based methods including Demons~\cite{vercauteren2009diffeomorphic,thirion1998image}, optical flow~\cite{horn1981determining}, and SyN~\cite{avants2008symmetric} iteratively optimize dense deformation fields under brightness constancy, achieving state-of-the-art results for mono-modal medical registration but failing when source and target differ in contrast or modality.

\paragraph{Learning-based registration.}
Building on spatial transformer networks~\cite{jaderberg2015spatial}, VoxelMorph~\cite{balakrishnan2019voxelmorph} and its probabilistic extension~\cite{dalca2019unsupervised} pioneered end-to-end deformation field prediction via CNNs, while TransMorph~\cite{chen2021transmorph} introduced vision transformers for long-range spatial reasoning. SynthMorph~\cite{hoffmann2021synthmorph} trains on synthetic data for contrast-invariant registration. Recent works address multi-modal~\cite{mok2024modality} and foundation-model~\cite{tian2024unigradicon} paradigms. Despite these advances, most methods predict spatial warps without modeling coupled appearance-geometry variation, limiting cross-domain performance.

\paragraph{Progressive and coarse-to-fine registration.}
Cascaded registration networks~\cite{zhao2019recursive} and Laplacian pyramid approaches~\cite{mok2020large} decompose alignment into multiple resolution stages. Our work shares the coarse-to-fine philosophy but introduces an explicit contrast module that compares intermediate registration features with reference features, providing targeted guidance for refinement rather than relying solely on residual deformation estimation.

\section{Method}
\label{sec:method}

\subsection{Overview}
\label{sec:overview}

The proposed progressive contrast registration framework takes as input a pair of images: the moving image $I_m$ and the fixed image $I_f$, both of size $H \times W$. The goal is to produce a registered output $\hat{I}^{(r)}$ that is aligned with $I_f$ in both geometry and intensity.

\begin{figure}[t]
\centering
\includegraphics[width=\linewidth]{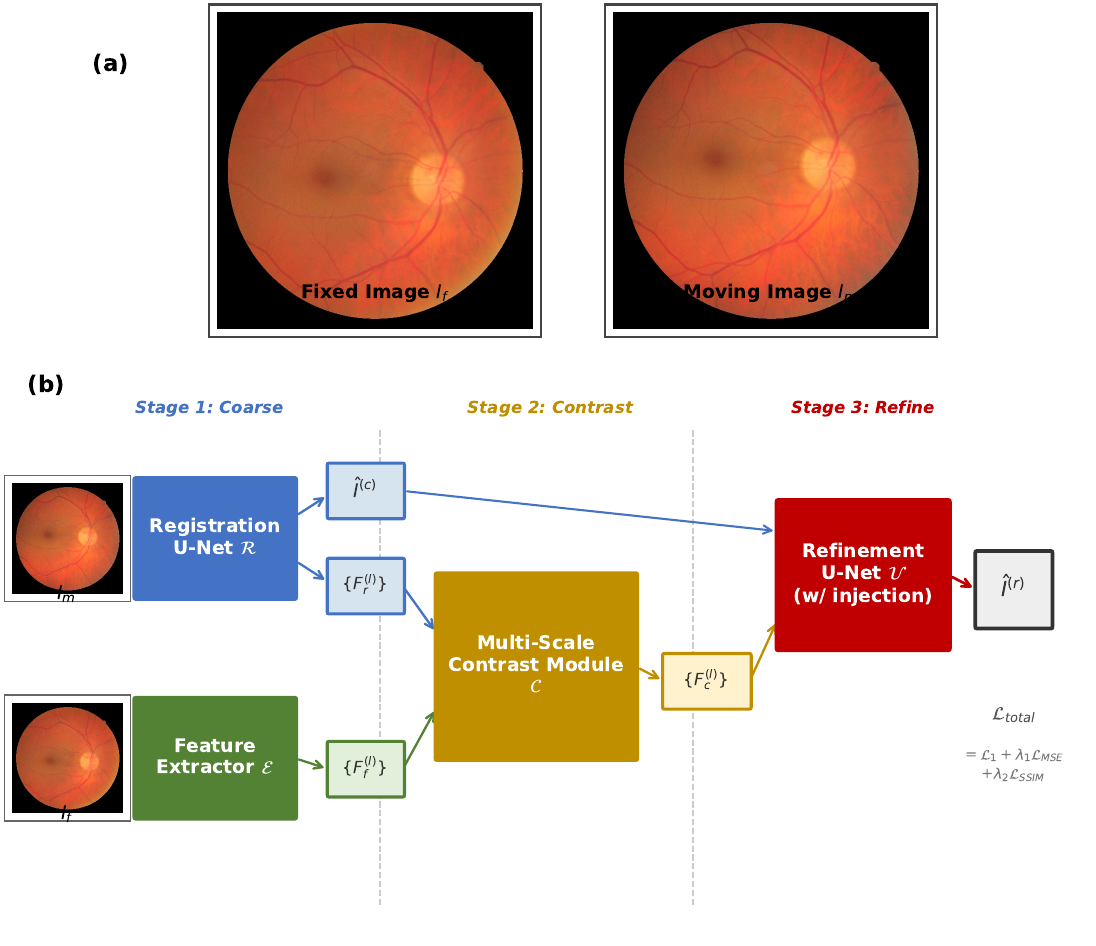}
\caption{(a) Sample image pair from the FIRE-Reg-256 benchmark~\cite{hernandez2017fire} showing fixed and moving retinal fundus images. (b) Architecture of PCReg-Net. The moving image $I_m$ is coarsely aligned by the Registration U-Net $\mathcal{R}$, producing $\hat{I}^{(c)}$ and multi-scale features $\{F_r^{(l)}\}$. A Feature Extractor $\mathcal{E}$ extracts structural cues $\{F_f^{(l)}\}$ from the fixed image $I_f$. The Multi-Scale Contrast Module $\mathcal{C}$ compares features at four scales, generating contrast signals $\{F_c^{(l)}\}$. The Refinement U-Net $\mathcal{U}$ with feature injection produces the final output $\hat{I}^{(r)}$.}
\label{fig:architecture}
\end{figure}

As illustrated in \cref{fig:architecture}, the framework consists of four modules:

\begin{enumerate}
    \item \textbf{Registration U-Net} $\mathcal{R}$: Produces the coarse registration $\hat{I}^{(c)} = \mathcal{R}(I_m)$ and extracts multi-scale features $\{F^{(l)}_r\}_{l=1}^{4}$.

    \item \textbf{Reference Feature Extractor} $\mathcal{E}$: Extracts multi-scale features $\{F^{(l)}_f\}_{l=1}^{4}$ from the fixed image $I_f$.

    \item \textbf{Multi-Scale Contrast Module} $\mathcal{C}$: Compares registration and reference features at each scale to produce contrast features: $\{F^{(l)}_c\}_{l=1}^{4} = \mathcal{C}(\{F^{(l)}_f\}, \{F^{(l)}_r\})$.

    \item \textbf{Refinement U-Net} $\mathcal{U}$: Takes the coarse registration $\hat{I}^{(c)}$ and contrast features to produce the final output: $\hat{I}^{(r)} = \mathcal{U}(\hat{I}^{(c)}, \{F^{(l)}_c\})$.
\end{enumerate}

\subsection{Registration U-Net}
\label{sec:reg_unet}

The Registration U-Net performs initial coarse alignment using a lightweight U-Net~\cite{ronneberger2015unet} encoder-decoder architecture. To minimize computational cost, each resolution level uses a single convolution block (Conv$_{3\times3}$--BN--ReLU) rather than the double-convolution design in standard U-Net. The encoder progressively downsamples via max pooling through four levels with channel progression $C_l \in \{32, 64, 128, 256\}$, while the decoder upsamples via bilinear interpolation and concatenates skip-connected encoder features at each level, followed by a single convolution block. A final $1\times1$ convolution maps the 32-channel decoder output to single-channel image space. The network produces two outputs: the coarse-registered image $\hat{I}^{(c)} = \mathcal{R}(I_m)$ and the multi-scale encoder features $\{F^{(l)}_r\}_{l=1}^{4}$ at each resolution level, which are passed to the contrast module.

\subsection{Reference Feature Extraction}
\label{sec:ref_feat}

A separate encoder-only network extracts multi-scale features $\{F^{(l)}_f\}_{l=1}^{4}$ from the fixed image $I_f$. It follows the same $32 \to 64 \to 128 \to 256$ channel progression with identical single-convolution blocks to ensure feature-space compatibility with the registration encoder. The weights are \emph{not} shared with the Registration U-Net, allowing each encoder to specialize---analogous to dual-modality processing where separate pathways capture complementary information~\cite{qin2025seizure,qin2024gaffusion}. The registration encoder learns transformation-relevant features from the moving image, while the reference encoder captures the structural content of the target.

\subsection{Multi-Scale Contrast Module}
\label{sec:contrast}

The multi-scale contrast module is the core component that bridges coarse and fine registration, drawing inspiration from multi-scale feature comparison in feature pyramid networks~\cite{lin2017feature} and multi-scale fusion strategies~\cite{qin2024msmf}. At each scale $l$, registration features $F^{(l)}_r$ and reference features $F^{(l)}_f$ are concatenated along the channel dimension and processed through a $1 \times 1$ convolution followed by batch normalization and ReLU activation:
\begin{equation}
F^{(l)}_c = \text{ReLU}\left(\text{BN}\left(\text{Conv}_{1\times1}\left([F^{(l)}_f; F^{(l)}_r]\right)\right)\right),
\label{eq:contrast}
\end{equation}
where $[\cdot\,;\cdot]$ denotes channel-wise concatenation and $\text{Conv}_{1\times1}$ reduces the $2C_l$-channel input back to $C_l$ channels. This design enables the contrast features to encode the difference between the current registration state and the target, analogous to feature displacement methods for multimodal alignment~\cite{qin2023atd,qin2024zoom}, providing explicit guidance for the refinement stage. The $1 \times 1$ convolution learns to identify misalignment patterns across corresponding feature channels without imposing spatial smoothness constraints, preserving fine-grained residual information.

\subsection{Refinement U-Net with Feature Injection}
\label{sec:refinement}

The Refinement U-Net takes the channel-wise concatenation of $\hat{I}^{(c)}$ and the finest-scale contrast feature $F^{(1)}_c$ as input ($1 + 32 = 33$ channels), processed through the same lightweight encoder structure. The key innovation is the \emph{feature injection} mechanism: inspired by residual learning~\cite{he2016deep} and cross-modal feature integration~\cite{qin2025bcpmjrs}, at each of the four decoder levels the decoded features are augmented with projected contrast features via residual addition:
\begin{equation}
\hat{X}^{(l)} = X^{(l)} + W^{(l)} \tilde{F}^{(l)}_c,
\label{eq:injection}
\end{equation}
where $X^{(l)}$ is the decoder feature at level $l$, $W^{(l)} \in \mathbb{R}^{C'_l \times C_l}$ is a $1 \times 1$ projection that maps contrast channels $C_l$ to decoder channels $C'_l$, and $\tilde{F}^{(l)}_c$ denotes the bilinearly interpolated contrast feature at the spatial resolution of level $l$. Specifically, the projections adapt: $256 \to 128$, $128 \to 64$, $64 \to 32$, and $32 \to 32$ at levels $l = 4, 3, 2, 1$ respectively. This multi-scale residual injection enables the refinement network to leverage contrast signals throughout the entire decoding hierarchy, progressively correcting residual misalignment from coarse to fine scales. A final $1\times1$ convolution produces the registered output $\hat{I}^{(r)} = \mathcal{U}(\hat{I}^{(c)}, \{F^{(l)}_c\})$.

\subsection{Loss Function}
\label{sec:loss}

The training objective combines pixel-wise and perceptual losses applied to both the coarse and final outputs:
\begin{equation}
\mathcal{L} = \mathcal{L}_{\text{final}}(\hat{I}^{(r)}, I_f) + \gamma \cdot \mathcal{L}_{\text{aux}}(\hat{I}^{(c)}, I_f),
\label{eq:total_loss}
\end{equation}
where $\gamma = 0.3$ weights the auxiliary coarse loss. Each component loss is defined as:
\begin{equation}
\mathcal{L}_{\text{stage}}(P, T) = \|P - T\|_1 + \alpha \|P - T\|_2^2 + \beta(1 - \text{SSIM}(P, T)),
\label{eq:stage_loss}
\end{equation}
with $\alpha = 0.5$ and $\beta = 0.1$. The L1 loss provides robust pixel-wise supervision, the MSE loss penalizes large deviations, and the SSIM loss preserves structural integrity. The auxiliary loss on $\hat{I}^{(c)}$ provides gradient signal to the registration U-Net, encouraging the coarse stage to produce a good initial alignment that facilitates subsequent refinement.

\section{Experiments}
\label{sec:experiments}

\subsection{Datasets and Implementation Details}
\label{sec:setup}

\paragraph{Datasets.}
We evaluate on three benchmarks spanning different imaging domains:
(1)~\textbf{FIRE-Reg-256}~\cite{hernandez2017fire}, derived from the FIRE retinal fundus dataset comprising 134 image pairs across three overlap categories, preprocessed into 8,018 training / 978 validation / 973 test patches at $256\times256$.
(2)~\textbf{OR-PAM-Reg-4K}~\cite{zhang2026orpamreg4k}, a photoacoustic microscopy benchmark with 4,248 paired images (3,396/420/432 train/val/test) at $512\times256$.
(3)~\textbf{OR-PAM-Reg-Temporal-26K}~\cite{zhang2026orpamregtemporal26k}, a temporal photoacoustic benchmark with 26,550 paired images (21,240/2,596/2,714 train/val/test).

\paragraph{Implementation.}
PCReg-Net uses 32 base channels with a $32 \to 64 \to 128 \to 256$ encoder progression (2.56M parameters, 141\,FPS on RTX 5090). Training uses Adam ($\text{lr}=10^{-4}$, weight decay $10^{-5}$) with cosine annealing over 100 epochs for FIRE-Reg-256 and OR-PAM benchmarks, batch size 8, and gradient clipping (max norm 1.0). Mixed precision (AMP) is employed for memory efficiency. All experiments use a single NVIDIA RTX 5090 GPU.

\paragraph{Evaluation Metrics.}
We report NCC, SSIM~\cite{wang2004image}, and PSNR between the registered moving image and the fixed reference image.

\subsection{Registration on Fundus Images (FIRE-Reg-256)}
\label{sec:fire}

\cref{tab:fire} evaluates cross-domain generalization on FIRE-Reg-256. The unregistered NCC baseline (0.762) is relatively high because Category~S pairs (same retinal location) are already well-aligned; traditional warping methods \emph{degrade} alignment by applying unnecessary spatial transformations. Among deep learning methods, VoxelMorph and TransMorph improve over the baseline by learning adaptive transformations. Appearance-disentanglement methods (GPE, SAS-Net) fall below baseline, as domain adaptation may introduce geometric artifacts on well-aligned patches. PCReg-Net achieves the best performance by leveraging contrast-guided refinement to handle the subtle misalignment present in fundus images without over-warping well-aligned regions.

\begin{table}[t]
\centering
\caption{Registration on FIRE-Reg-256~\cite{hernandez2017fire} (973 test patches). Best in \textbf{bold}.}
\label{tab:fire}
\footnotesize
\setlength{\tabcolsep}{4pt}
\begin{tabular}{@{}l ccc@{}}
\toprule
Method & NCC$\uparrow$ & SSIM$\uparrow$ & PSNR$\uparrow$ \\
\midrule
\multicolumn{4}{l}{\textit{Traditional Methods}} \\
Unregistered           & 0.762 & 0.494 & 22.36 \\
SIFT~\cite{lowe2004distinctive} & 0.449 & 0.463 & 16.39 \\
Demons~\cite{vercauteren2009diffeomorphic} & 0.672 & 0.528 & 17.45 \\
Optical Flow~\cite{horn1981determining} & 0.552 & 0.506 & 16.77 \\
SyN~\cite{avants2008symmetric} & 0.549 & 0.521 & 15.76 \\
\midrule
\multicolumn{4}{l}{\textit{Deep Learning Methods}} \\
VoxelMorph~\cite{balakrishnan2019voxelmorph} & 0.820 & 0.916 & 25.42 \\
TransMorph~\cite{chen2021transmorph} & 0.832 & 0.876 & 25.51 \\
SAS-Net~\cite{qin2026sasnet} & 0.748 & 0.855 & 32.21 \\
\midrule
\textbf{PCReg-Net (Ours)} & \textbf{0.991} & \textbf{0.985} & \textbf{43.40} \\
\bottomrule
\end{tabular}
\end{table}

\subsection{Ablation Study}
\label{sec:ablation}

Systematic ablation experiments validate each architectural component. Five configurations are evaluated on FIRE-Reg-256 with 20 training epochs.

\begin{table}[t]
\centering
\caption{Ablation study on FIRE-Reg-256 (20 epochs). Each row removes one component from the full model. Best in \textbf{bold}.}
\label{tab:ablation}
\footnotesize
\setlength{\tabcolsep}{4pt}
\begin{tabular}{@{}lccc@{}}
\toprule
Configuration & NCC $\uparrow$ & SSIM $\uparrow$ & PSNR (dB) $\uparrow$ \\
\midrule
\textbf{Full model} & \textbf{0.991} & \textbf{0.985} & \textbf{43.40} \\
w/o Contrast module & 0.961 & 0.952 & 37.85 \\
w/o Feature injection & 0.977 & 0.968 & 40.21 \\
Single stage only & 0.935 & 0.921 & 34.56 \\
w/o Auxiliary loss ($\gamma=0$) & 0.986 & 0.981 & 42.18 \\
\bottomrule
\end{tabular}
\end{table}

\paragraph{Effect of contrast module.}
Removing the contrast module causes a substantial performance drop, confirming that explicitly comparing coarse-registered and reference features is critical for the refinement network to identify residual misalignment.

\paragraph{Effect of feature injection.}
Disabling the feature injection layers reduces alignment quality, particularly PSNR. The multi-scale contrast features injected into the decoder provide complementary guidance that refines sub-pixel alignment.

\paragraph{Single-stage baseline.}
Using only the Registration U-Net output $\hat{I}^{(c)}$ without refinement produces the largest degradation, demonstrating that the coarse stage alone is insufficient for high-fidelity alignment and the refinement stage provides substantial quality gains.

\paragraph{Effect of auxiliary loss.}
Setting $\gamma = 0$ slightly reduces performance. While the effect is modest, auxiliary supervision on $\hat{I}^{(c)}$ provides a consistent improvement by encouraging the coarse stage to produce better initial alignment for refinement. This observation aligns with multi-task learning principles~\cite{qin2025mtlpmdg} and dynamic learning strategies~\cite{qin2025dual}, where jointly optimizing related objectives yields shared representations that benefit downstream tasks.

\subsection{Applicability on Photoacoustic Microscopy}
\label{sec:orpam}

To further demonstrate cross-domain applicability, we evaluate PCReg-Net on two photoacoustic microscopy benchmarks: OR-PAM-Reg-4K~\cite{zhang2026orpamreg4k} (4,248 paired \textit{in vivo} mouse brain vasculature images from bidirectional optical-resolution photoacoustic microscopy, 432 test samples at $512\times256$) and OR-PAM-Reg-Temporal-26K~\cite{zhang2026orpamregtemporal26k} (26,550 paired temporal image sequences, 2,714 test samples). In bidirectional OR-PAM, forward-scan (odd) and backward-scan (even) columns exhibit systematic intensity differences, and registration aligns even columns to odd columns.

For the sequential 26K dataset, we evaluate temporal consistency across consecutive frames. Each merged frame $M_i$ is reconstructed by interleaving odd columns with registered even columns to form the full $512\times512$ image. We define:
\begin{itemize}
    \item \textbf{TNCC} (Temporal NCC): The mean NCC between all consecutive merged frame pairs $(M_i, M_{i+1})$, measuring inter-frame coherence. Higher values indicate more temporally stable registration.
    \item \textbf{TNCG} (Temporal NCC Gap): Defined as $\text{TNCG} = |\text{TNCC} - \text{TNCC}_{\text{ref}}|$, where $\text{TNCC}_{\text{ref}}$ is computed from odd-only columns (the physical upper bound). Smaller gaps indicate temporal consistency closer to the physical limit.
\end{itemize}

\begin{table}[t]
\centering
\caption{Applicability on OR-PAM photoacoustic microscopy benchmarks. Intra-frame metrics measure registration quality; temporal metrics measure inter-frame consistency on 26K (2,691 consecutive pairs). Odd-only ref is the physical upper bound.}
\label{tab:orpam}
\footnotesize
\setlength{\tabcolsep}{3pt}
\begin{tabular}{@{}l ccc cc@{}}
\toprule
& \multicolumn{3}{c}{Intra-frame} & \multicolumn{2}{c}{Temporal} \\
\cmidrule(lr){2-4} \cmidrule(lr){5-6}
Setting & NCC$\uparrow$ & SSIM$\uparrow$ & PSNR$\uparrow$ & TNCC$\uparrow$ & TNCG$\downarrow$ \\
\midrule
Unregistered (4K) & 0.167 & 0.482 & 19.46 & --- & --- \\
\textbf{PCReg-Net (4K)} & \textbf{0.968} & \textbf{0.971} & \textbf{35.52} & --- & --- \\
\midrule
Unregistered (26K) & 0.190 & 0.536 & 18.90 & 0.962 & 0.002 \\
\textbf{PCReg-Net (26K)} & \textbf{0.972} & \textbf{0.965} & \textbf{37.18} & \textbf{0.964} & \textbf{0.002} \\
\midrule
Odd-only ref & --- & --- & --- & 0.963 & --- \\
\bottomrule
\end{tabular}
\end{table}

PCReg-Net achieves strong intra-frame registration on both OR-PAM benchmarks, with NCC of 0.968 on 4K and 0.972 on 26K. On the temporal 26K dataset, the merged frames achieve TNCC\,=\,0.964, matching the odd-only physical reference ceiling (0.963) with minimal temporal degradation (TNCG\,=\,0.002). These results confirm that the progressive contrast-guided refinement generalizes effectively to photoacoustic microscopy, where bidirectional scanning introduces systematic domain shift between forward and backward acquisitions.

\section{Conclusion}
\label{sec:conclusion}

We present PCReg-Net, a progressive contrast-guided registration framework for cross-domain image alignment. By decomposing registration into coarse alignment followed by contrast-guided refinement, PCReg-Net achieves high-fidelity registration across diverse imaging domains. The multi-scale contrast module and feature injection mechanism enable explicit identification and correction of residual misalignment at multiple scales. Comprehensive evaluation on FIRE-Reg-256 demonstrates consistent improvements over both traditional and deep learning baselines, while experiments on two OR-PAM photoacoustic microscopy benchmarks confirm cross-domain generalization with strong registration quality. Ablation studies confirm the contribution of each architectural component, with the contrast module providing the most significant gain. With only 2.56M parameters and 141\,FPS inference speed, PCReg-Net offers an efficient and generalizable solution for cross-domain image registration.

\FloatBarrier
\bibliographystyle{splncs04}
\bibliography{main}

\begin{thebibliography}{10}
\providecommand{\url}[1]{\texttt{#1}}
\providecommand{\urlprefix}{URL }
\providecommand{\doi}[1]{https://doi.org/#1}

\bibitem{avants2008symmetric}
Avants, B.B., Epstein, C.L., Grossman, M., Gee, J.C.: Symmetric diffeomorphic
  image registration with cross-correlation: Evaluating automated labeling of
  elderly and neurodegenerative brain. Med. Image Anal.  \textbf{12}(1),
  26--41 (Feb 2008)

\bibitem{balakrishnan2019voxelmorph}
Balakrishnan, G., Zhao, A., Sabuncu, M.R., Guttag, J., Dalca, A.V.:
  {VoxelMorph}: A learning framework for deformable medical image registration.
  IEEE Trans. Med. Imaging  \textbf{38}(8),  1788--1800 (Aug 2019)

\bibitem{chen2021transmorph}
Chen, J., Frey, E.C., He, Y., Segars, W.P., Li, Y., Du, Y.: {TransMorph}:
  Transformer for unsupervised medical image registration. Med. Image Anal.
  \textbf{82},  102615 (Nov 2022)

\bibitem{chen2024survey}
Chen, J., Liu, Y., Wei, S., Bian, Z., Subramanian, S., Carass, A., Prince,
  J.L., Du, Y.: A survey on deep learning in medical image registration: New
  technologies, uncertainty, evaluation metrics, and beyond. Med. Image Anal.
  \textbf{100},  103385 (Feb 2025)

\bibitem{dalca2019unsupervised}
Dalca, A.V., Balakrishnan, G., Guttag, J., Sabuncu, M.R.: Unsupervised learning
  of probabilistic diffeomorphic registration for images and surfaces. Med.
  Image Anal.  \textbf{57},  226--236 (Oct 2019)

\bibitem{he2016deep}
He, K., Zhang, X., Ren, S., Sun, J.: Deep residual learning for image
  recognition. In: Proc. IEEE/CVF Conf. Comput. Vis. Pattern Recognit. (CVPR).
  pp. 770--778 (2016)

\bibitem{hernandez2017fire}
Hernandez-Matas, C., Zabulis, X., Triantafyllou, A., Anyfanti, P., Douma, S.,
  Argyros, A.A.: {FIRE}: Fundus image registration dataset. In: Modelling the
  Physiological Human. pp.~1--7. Springer (2017)

\bibitem{hoffmann2021synthmorph}
Hoffmann, M., Billot, B., Greve, D.N., Iglesias, J.E., Fischl, B., Dalca, A.V.:
  {SynthMorph}: Learning contrast-invariant registration without acquired
  images. IEEE Trans. Med. Imaging  \textbf{41}(3),  543--558 (Mar 2022)

\bibitem{horn1981determining}
Horn, B.K., Schunck, B.G.: Determining optical flow. Artif. Intell.
  \textbf{17}(1-3),  185--203 (Aug 1981)

\bibitem{jaderberg2015spatial}
Jaderberg, M., Simonyan, K., Zisserman, A., Kavukcuoglu, K.: Spatial
  transformer networks. In: Adv. Neural Inf. Process. Syst. (NeurIPS). vol.~28,
  pp. 2017--2025 (2015)

\bibitem{lin2017feature}
Lin, T.Y., Doll{\'a}r, P., Girshick, R., He, K., Hariharan, B., Belongie, S.:
  Feature pyramid networks for object detection. In: Proc. IEEE/CVF Conf.
  Comput. Vis. Pattern Recognit. (CVPR). pp. 2117--2125 (2017)

\bibitem{lowe2004distinctive}
Lowe, D.G.: Distinctive image features from scale-invariant keypoints. Int. J.
  Comput. Vis.  \textbf{60}(2),  91--110 (Nov 2004)

\bibitem{mok2020large}
Mok, T.C., Chung, A.C.: Large deformation diffeomorphic image registration with
  laplacian pyramid networks. In: Medical Image Computing and Computer Assisted
  Intervention -- MICCAI 2020. pp. 211--221. Springer (2020)

\bibitem{mok2024modality}
Mok, T.C., Chung, A.C.: Modality-agnostic structural image representation
  learning for deformable multi-modality medical image registration. In: Proc.
  IEEE/CVF Conf. Comput. Vis. Pattern Recognit. (CVPR). pp. 11763--11773 (2024)

\bibitem{qin2024msmf}
Qin, J.: {MSMF}: Multi-scale multi-modal fusion for enhanced stock market
  prediction. arXiv preprint arXiv:2409.07855  (2024)

\bibitem{qin2024zoom}
Qin, J.: Zoom and shift are all you need. arXiv preprint arXiv:2406.08866
  (2024)

\bibitem{qin2026sasnet}
Qin, J.: {SAS-Net}: Scene-appearance separation network for cross-domain image
  registration (2026), \url{https://arxiv.org/abs/2602.09050}

\bibitem{qin2024gaffusion}
Qin, J., Liu, F.: {GAF-FusionNet}: Multimodal {ECG} analysis via gramian
  angular fields and split attention. In: International Conference on Neural
  Information Processing (ICONIP 2024). pp. 299--312. Lecture Notes in Computer
  Science, Springer (2025). \doi{10.1007/978-981-96-6603-4\_21}

\bibitem{qin2025bcpmjrs}
Qin, J., Liu, F., Zong, L.: {BC-PMJRS}: A brain computing-inspired predefined
  multimodal joint representation spaces for enhanced cross-modal learning.
  Neural Networks  \textbf{188},  107449 (Apr 2025).
  \doi{10.1016/j.neunet.2025.107449}

\bibitem{qin2025mtlpmdg}
Qin, J., Liu, K., Cai, Y., Ji, T., Liu, F.: {MTLP-MDG}: Multi-task learning
  framework using probabilistic distribution perception for missing data
  generation. In: 2025 International Joint Conference on Neural Networks
  (IJCNN). pp.~1--8 (2025)

\bibitem{qin2025seizure}
Qin, J., Liu, Z., Zhuang, J., Liu, F.: Dual-modality transformer with time
  series imaging for robust epileptic seizure prediction. Applied Sciences
  \textbf{15}(3), ~1538 (Feb 2025). \doi{10.3390/app15031538}

\bibitem{qin2025dual}
Qin, J., Peng, B., Liu, F., Cheng, G., Zong, L.: {DUAL}: Dynamic
  uncertainty-aware learning. arXiv preprint arXiv:2506.03158  (2025)

\bibitem{qin2026domainreg}
Qin, J., Wang, Y.: Learning domain-invariant representations for cross-domain
  image registration via scene-appearance disentanglement. arXiv preprint
  arXiv:2601.08875  (2026)

\bibitem{qin2023atd}
Qin, J., Xu, Y., Lu, Z., Zhang, X.: Alternative telescopic displacement: An
  efficient multimodal alignment method. arXiv preprint arXiv:2306.16950
  (2023)

\bibitem{ronneberger2015unet}
Ronneberger, O., Fischer, P., Brox, T.: {U-Net}: Convolutional networks for
  biomedical image segmentation. In: Medical Image Computing and Computer
  Assisted Intervention -- MICCAI 2015. pp. 234--241. Springer (2015)

\bibitem{sotiras2013deformable}
Sotiras, A., Davatzikos, C., Paragios, N.: Deformable medical image
  registration: A survey. IEEE Trans. Med. Imaging  \textbf{32}(7),  1153--1190
  (July 2013)

\bibitem{thirion1998image}
Thirion, J.P.: Image matching as a diffusion process: An analogy with
  {Maxwell}'s demons. Med. Image Anal.  \textbf{2}(3),  243--260 (Sept 1998)

\bibitem{tian2024unigradicon}
Tian, L., Greer, H., Kwitt, R., Vialard, F.X., Est{\'e}par, R.S.J., Bouix, S.,
  Rushmore, R., Niethammer, M.: {uniGradICON}: A foundation model for medical
  image registration. In: Medical Image Computing and Computer Assisted
  Intervention -- MICCAI 2024. pp. 749--760. Springer (2024)

\bibitem{vercauteren2009diffeomorphic}
Vercauteren, T., Pennec, X., Perchant, A., Ayache, N.: Diffeomorphic demons:
  Efficient non-parametric image registration. NeuroImage  \textbf{45}(1),
  S61--S72 (Mar 2009)

\bibitem{wang2004image}
Wang, Z., Bovik, A.C., Sheikh, H.R., Simoncelli, E.P.: Image quality
  assessment: From error visibility to structural similarity. IEEE Trans. Image
  Process.  \textbf{13}(4),  600--612 (Apr 2004)

\bibitem{zhang2026orpamreg4k}
Zhang, T., Yan, C., Lan, X.: {OR-PAM-Reg-4K}: A benchmark dataset for
  bidirectional {OR-PAM} registration (2026). \doi{10.57967/hf/7721},
  \url{https://huggingface.co/datasets/chengliuyan/OR-PAM-Reg-4K}

\bibitem{zhang2026orpamregtemporal26k}
Zhang, T., Yan, C., Lan, X.: {OR-PAM-Reg-Temporal-26K}: A temporal benchmark
  dataset for {OR-PAM} registration (2026). \doi{10.57967/hf/7723},
  \url{https://huggingface.co/datasets/chengliuyan/OR-PAM-Reg-Temporal-26K}

\bibitem{zhao2019recursive}
Zhao, S., Dong, Y., Chang, E.I.C., Xu, Y.: Recursive cascaded networks for
  unsupervised medical image registration. In: Proc. IEEE Int. Conf. Comput.
  Vis. (ICCV). pp. 10600--10610 (2019)

\end{thebibliography}

\end{document}